\documentclass[10pt,twocolumn,letterpaper]{article}

\usepackage[pagenumbers]{cvpr} 

\usepackage[accsupp]{axessibility} 
\usepackage{multirow} 
\usepackage{xcolor}
\usepackage{amsmath} 
\usepackage{graphicx} 
\usepackage{booktabs} 
\usepackage{bbding} 
\usepackage{wrapfig} 
\usepackage{colortbl} 
\usepackage{caption} 
\usepackage{array} 
\usepackage{diagbox} 
\usepackage[table]{xcolor}
\usepackage{booktabs}       
\usepackage{amsfonts}       
\usepackage{microtype}      
\usepackage{tabularx} 
\usepackage{pifont} 
\usepackage{cuted}

\definecolor{cvprblue}{rgb}{0.21,0.49,0.74}
\usepackage[pagebackref,breaklinks,colorlinks,allcolors=cvprblue]{hyperref}

\def\paperID{*****} 
\def\confName{CVPR}
\def\confYear{2026}

\title{SocialMirror: Reconstructing 3D Human Interaction Behaviors from Monocular Videos with Semantic and Geometric Guidance}

\author{
Qi Xia$^1$ \quad
Peishan Cong$^1$  \quad
Ziyi Wang$^1$  \quad
Yujing Sun$^2$  \quad
Qin Sun$^{3,4}$  \\
Xinge Zhu$^5$  \quad
Mao Ye$^6$  \quad
Ruigang Yang$^7$  \quad
Yuexin Ma$^{1,}$ \thanks{Corresponding author. This work was supported by MoE Key Laboratory of Intelligent Perception and Human-Machine Collaboration (KLIP-HuMaCo), Shanghai Frontiers Science Center of Human-centered Artificial Intelligence (ShangHAI).} \\
\\
$^{1}$ ShanghaiTech University \quad
$^{2}$ Nanyang Technological University \\
$^{3}$ Guangzhou Institute of Energy Conversion, CAS \quad
$^{4}$ University of Science and Technology of China \\
$^{5}$ The Chinese University of Hong Kong \quad
$^{6}$ Inceptio Technology \quad 
$^{7}$ Shanghai Jiao Tong University
}

\begin{document}
\maketitle

\begin{strip}
    \centering
    \vspace{-14ex}
    \includegraphics[width=\linewidth]{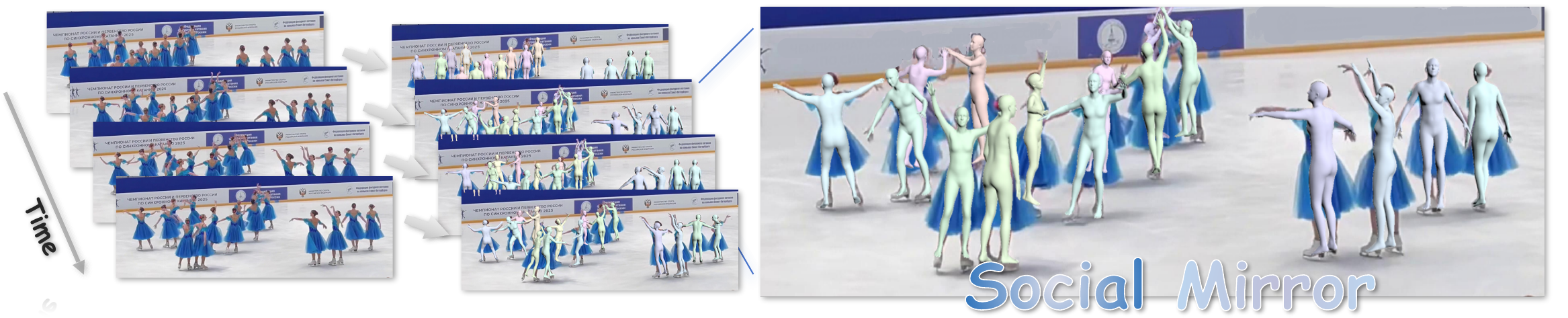}
    \vspace{-2ex}
    \captionof{figure}{We reconstruct 3D human motion from monocular videos, specifically targeting close interaction scenarios. By leveraging both semantic and geometric guidance, SocialMirror resolves ambiguities through infilling and ensuring the spatial relationships.}
    \label{fig:teasor}
\end{strip}

\begin{abstract}

Accurately reconstructing human behavior in close-interaction scenarios is crucial for enabling realistic virtual interactions in augmented reality, precise motion analysis in sports, and natural collaborative behavior in human-robot tasks. Reliable reconstruction in these contexts significantly enhances the realism and effectiveness of AI-driven interactive applications.
However, human reconstruction from monocular videos in close-interaction scenarios remains challenging due to severe mutual occlusions, which introduce local motion ambiguity, disrupted temporal continuity and erroneous spatial relationships. In this paper, we propose SocialMirror, a diffusion-based framework that integrates semantic and geometric cues to effectively address these issues. 
Specifically, we first leverage high-level interaction descriptions generated by a vision-language model to guide a semantic-guided motion infiller, which infers occluded body poses and resolves local pose ambiguities. 
Next, we propose a sequence-level Temporal Motion Refiner that enforces smooth, jitter-free motions, while incorporating geometric constraints during sampling to ensure plausible contact and spatial relationships.
Evaluations on multiple interaction benchmarks show that SocialMirror achieves state-of-the-art performance in reconstructing interactive human meshes, demonstrating strong generalization across unseen datasets and in-the-wild scenarios. 

\end{abstract}    
\section{Introduction}

\label{sec:introduction}

Human reconstruction, which recovers the 3D geometry and motion of human bodies from visual inputs, is a fundamental computer vision task with extensive applications in fields such as augmented reality~\cite{urgo2024ai}, sports analysis~\cite{fukushima2024potential, xi2024enhancing}, and film animation.
Close human interactions~\cite{huang2024closeInt,muller2024buddi}, including social and competitive behaviors, are particularly critical in these contexts. Such interactions further play a crucial role in robotics applications, where collaborative tasks require seamless human-robot interaction. Accurately modeling human behavior in such interactions allows robots to engage in more natural, human-like collaborations, aligning with human preferences and enhancing the effectiveness of AI in interactive tasks.

Previous monocular human reconstruction works~\cite{kanazawa2018end,bogo2016keep} primarily target single-person scenarios. These methods typically focus on accurate pose estimation~\cite{li2021hybrik, rempe2021humor}, shape reconstruction fidelity~\cite{goel2023humans4D, pavlakos2019expressive, xu2020ghum}, or temporal smoothness across frames~\cite{kocabas2020vibe, zheng20213sttrans, zeng2022smoothnet}. However, limiting reconstruction to single-person scenarios restricts applicability in real-world multi-person interactive settings. 
A few works~\cite{Huang2023ReconstructingGO,lu2023dposer,ugrinovic2024multiphys,sun2022BEV, sat-hmr2025, newell2025comotion, Liu_2025_CVPR} have considered reconstructing multi-human poses, employing techniques such as explicit collision avoidance constraints~\cite{ugrinovic2024multiphys}, depth ordering modeling in crowded scenes~\cite{sun2022BEV, wen2023crowd3d}, data-driven priors~\cite{zhu2024dpmesh, lu2023dposer, rempe2021humor, shi2023phasemp}, or relational reasoning~\cite{Huang2023ReconstructingGO}. 
The methods mentioned above typically address multi-person scenarios, where individuals are in the same space but not directly interacting. In contrast, close-interaction scenes often involve heavy occlusions, especially when individuals are physically touching or positioned in tight spaces, which is more relevant for collaborative tasks and robot-human interactions. 
While some methods~\cite{muller2024buddi,huang2024closeInt,fang2024capturing} have used mutual priors to model interactions, reconstructing closely interacting humans from monocular videos remains challenging due to the exclusive reliance on image features under severe occlusions.
This leads to three critical issues:
 (1) \textbf{Local pose ambiguities} occur when one person occludes another, so the pose of the hidden body cannot be reliably inferred from pixels alone.
 (2) \textbf{Temporal inconsistencies} arise when occlusions disrupt tracking continuity, resulting in unrealistic motion, such as sudden changes in pose.
 (3) \textbf{Spatial relationship errors} happen when image features alone fail to capture the dynamic, complex interactions between individuals in close proximity, causing inaccuracies in contact areas.

We observe that human interactions are inherently intentional, suggesting the necessity of incorporating semantic context to infer motion and spatial relationships. Additionally, as interactions naturally occur within 3D space, enforcing geometric constraints is crucial for achieving physically plausible reconstructions. Motivated by these insights, we propose \textbf{SocialMirror}, a Semantic and Geometric guided framework for Interactive Human Mesh Reconstruction from monocular video.
First, we introduce \textbf{\textit{Semantic-Guided Motion Infiller}} which incorporates textual semantic guidance alongside visual features to recover motion in occluded regions, effectively addressing severe occlusions and image-feature degradation. Specifically, a Vision-Language Model (VLM)~\cite{bai2025qwen2,achiam2023gpt} Annotator first generates textual descriptions of human interactions, providing essential semantic context and temporal cues. These annotations, along with image features from visible body parts extracted by a pre-trained backbone, guide the process to reconstruct infilled motion from invisible regions, ensuring semantically coherent reconstruction despite occlusions.
To model spatial relationships between closely interacting people, the \textbf{\textit{Geometry Optimizer}} encodes geometric structure from 3D joint positions and produces guidance signals for contact-aware refinement.
The \textbf{\textit{Temporal Motion Refiner}} then performs sequence-level optimization, merging diffusion-based infilled motion with reconstructions from visible regions while incorporating Geometry Optimizer guidance. This yields spatially and temporally coherent results even when image information is severely degraded or missing.

Experimental evaluations demonstrate that the proposed method achieves superior reconstruction accuracy on human interaction datasets with particular advantages in capturing interpersonal spatial relationships and interaction plausibility. To summarize, our work makes the following contributions:
 (1) We introduce semantic information into monocular video-based human mesh reconstruction via a diffusion-based framework. Semantic guidance enables the network to infer plausible poses in occluded regions, effectively resolving ambiguities through motion infilling. In addition, semantic context provides essential temporal cues and contact relationships, enhancing reconstruction accuracy in closely interacting regions.
 (2) The proposed method combines 3D geometric guidance with temporal refinement, encouraging consistent spatial relations, smooth motion, and physically plausible contact.
 (3) Experimental validation confirms our method achieves superior reconstruction quality in monocular interactive human scenarios. Notably, the approach demonstrates generalization capabilities across unseen datasets and in-the-wild scenarios.

\section{Related Work}

\subsection{Human reconstruction}

Building on advancements in single-person 3D reconstruction~\cite{kanazawa2018end,bogo2016keep}, recent approaches have increasingly focused on joint reconstruction of multiple individuals from monocular images. 
Prior works~\cite{fieraru2021remips, jiang2020coherent, zanfir2018monocular, sun2021ROMP, sun2022BEV, fieraru2020CHI3D, Li2022CLIFFCL} have focused on improving human relative position and depth estimation, using strategies such as depth ordering losses~\cite{fieraru2021remips, jiang2020coherent}, collision constraints~\cite{zanfir2018monocular, sun2021ROMP}, and bird's-eye view depth reasoning~\cite{sun2022BEV}. However, these methods still struggle with occlusions and the modeling of interpersonal relationships. 
To address these issues, some works enhance feature extraction under occlusion~\cite{kocabas2021pare, multi-hmr2024}, incorporate pose priors~\cite{zhu2024dpmesh, lu2023dposer, rempe2021humor, shi2023phasemp}, or use contextual motion completion frameworks~\cite{yuan2022glamr} to infill unseen human motions. Additionally, GroupRec~\cite{Huang2023ReconstructingGO} improves human mesh recovery through relational reasoning. 
However, these approaches fail to capture the complex interpersonal interactions and delicate contact in close-range scenarios.
Only a few studies~\cite{muller2024buddi, ugrinovic2024multiphys, huang2024closeInt,fang2024capturing} explicitly address close interactions, which involve more intimate contact and heavy occlusions. BUDDI~\cite{muller2024buddi} introduces a diffusion-based prior but is limited to static images. MultiPhys~\cite{ugrinovic2024multiphys} resolves mesh interpenetration using a physics engine, while CloseInt~\cite{huang2024closeInt} applies mutual attention modules for iterative refinement from monocular video. Nevertheless, all of these methods overlook the semantic context inherent in close human interactions and still face challenges with visual ambiguities.

\subsection{Human motion generation}

Human motion generation has progressed from single-person motion generation~\cite{mdm2022human,mld,jiang2024motiongpt} to more complex human-human interaction generation. Approaches include response synthesis~\cite{chopin2023interaction, ghosh2024remos, liu2023interactive, xu2024regennet}, where motion is generated in response to an actor's movements, and interaction generation~\cite{liang2024intergen, tanaka2023role}, which generates motions for all interacting individuals simultaneously. 
The methods mentioned above primarily focus on motion generation without explicit control. Control-based approaches, such as motion completion~\cite{choi2021ilvr, chung2022improving, zhao2024dart, mdm2022human} and trajectory- or joint-based~\cite{wan2024tlcontrol} control frameworks, have introduced finer control, improving the coherence and diversity of the generated motion. OmniControl~\cite{xie2023omnicontrol} and InterControl~\cite{wang2023intercontrol} integrate ControlNet~\cite{zhang2023adding} to enforce joint constraints and physical plausibility, ensuring more accurate and realistic motion.
Control-based diffusion is especially beneficial for human reconstruction tasks, where accurate alignment with input images and the ability to handle occlusions or depth ambiguities are crucial for maintaining both temporal and spatial consistency in the generated motions.

\subsection{LLM in pose estimation}

Large language models (LLMs)~\cite{achiam2023gpt} are known for their strong generalization capabilities, particularly in introducing semantic information. Their semantic flexibility and generalization have been proven effective in pose estimation tasks. Xiao et al.~\cite{xiao2025occluded}, for instance, integrates image features with CLIP text-image embeddings, creating multimodal conditional inputs that improve pose understanding. 
PromptHMR~\cite{prompt-hmr2025} uses SHAPY~\cite{Shapy:CVPR:2022} to generate body shape description texts and fuses the encoded text prompts with other features to improve monocular body shape estimation accuracy.
Subramanian et al.~\cite{subramanian2024pose} leverage a LLM to generate contact constraints between body parts, transforming these into a loss function to enforce physically consistent predictions for both self-contact and interpersonal interactions. Xu et al.~\cite{xu2025adapting} use a vision-language model (VLM) to extract detailed descriptions of body part interactions, which are then used as multimodal feedback to refine initial pose estimates. 
Building on this, we extend these techniques to human reconstruction from monocular videos in close-interaction scenarios, where we combine visual cues from input images and textual cues from VLM~\cite{bai2025qwen2} to generate more accurate and semantically coherent motion. Furthermore, we incorporate temporal contact labels and refine the reconstruction process to ensure not only spatial consistency but also temporal continuity and geometric plausibility.
\section{Method}

\begin{figure*}
    \vspace{-4ex}
    \centering
    \includegraphics[width=0.98\linewidth]{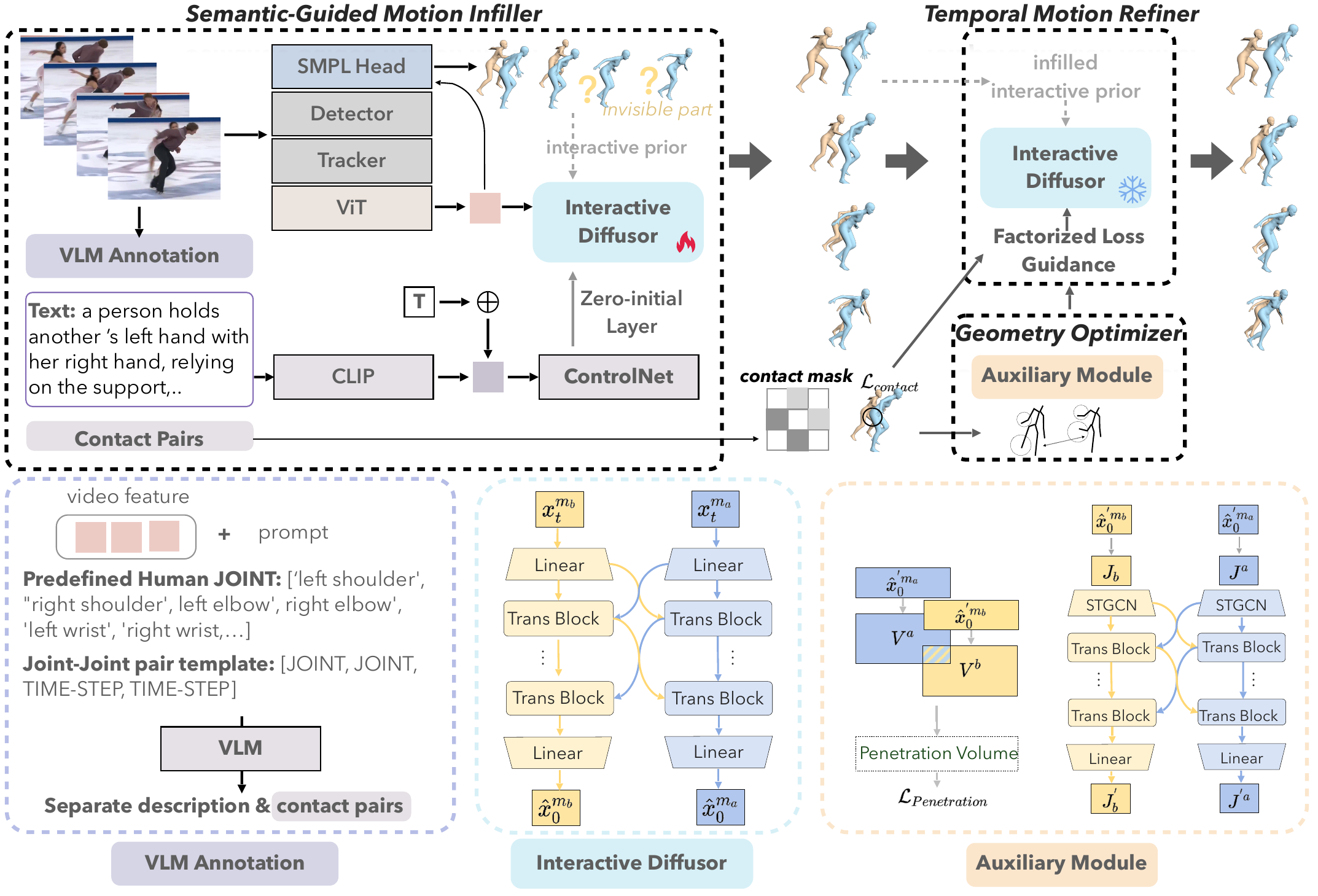}
    \vspace{-2ex}
    \caption{The framework of SocialMirror, which integrates semantic guidance from vision-language annotations and further refine the result with geometric constraints. \textit{Trans Block} refers to the transformer block.}
    \label{fig:pipeline}
    \vspace{-2ex}
\end{figure*}

The aim of this work is to reconstruct human close interactions from monocular videos. We introduce SocialMirror, a semantic and geometry-guided diffusion-based framework for interactive human mesh reconstruction, as shown in Figure~\ref{fig:pipeline}. Specifically, we extract textual descriptions and labels with temporal and close-contact information from a VLM and integrate these semantic features with visual data in Semantic-Guided Motion Infiller~\ref{sec:sem}, which compensates for visual feature degradation under severe occlusions and mitigates local pose ambiguities. The Geometry Optimizer~\ref{sec:geo} uses an auxiliary model to optimize 3D joint positions, generating geometric guidance signals to better model spatial relationships. The Temporal Motion Refiner~\ref{sec:tem} refines the reconstruction results based on these geometric signals, ensuring temporal consistency.

\subsection{Semantic-Guided Motion Infiller}
\label{sec:sem}

In multi-person close-interaction scenarios, severe partial occlusion frequently occurs, causing certain individuals to become visually obscured. Under such challenging conditions, existing reconstruction methods~\cite{muller2024buddi,huang2024closeInt} typically struggle due to the lack of reliable visual features from occluded subjects. Nevertheless, human observers consistently maintain perceptual coherence in these scenarios by effectively utilizing semantic information: even when visual details are obscured, contextual cues allow humans to infer plausible states of hidden regions via spatial and temporal reasoning. Inspired by this observation, we argue that motion reconstruction should leverage high-level semantics rather than relying solely on pixel-level cues. Consequently, our aim is to enable models to learn semantic-to-motion mappings, empowering the model to inpaint invisible regions through available visual cues and inferred interaction semantics.

\noindent \textbf{VLM Annotator.}
Large language models offer strong generalization capabilities and rich semantic information. Leveraging inputs such as detailed background scene data, human joint information, and predefined instructions, we use a vision-language model to generate semantic captions for the interactive motion of two people. These captions are then converted into single-person descriptions through prompt engineering. Additionally, we introduce sequential and spatial-level contact labels to guide the language model in modeling interactions, which are used in the Temporal Motion Refiner and Geometry Optimizer. We calculate the minimum distance between joints of the individuals and label pairs with a distance below a threshold as contact. Each contact pair is then formatted as (JOINT, JOINT, BEGIN-CONTACT-TIME-STEP, END-CONTACT-TIME-STEP) for further processing. We fine-tune the VLM to enable the model to infer contact labels. Details of the template design are provided in the Appendix.

\noindent \textbf{Feature Extractor.}
The input is an image sequence of length $L$, and the output is the SMPL parameters for each person describing their motion: local pose $\theta \in \mathbb{R}^{21 \times 3}$, shape $\beta \in \mathbb{R}^{10}$, rotation $\phi \in \mathbb{R}^{3}$, and translation $\tau \in \mathbb{R}^{3}$. The parameters for a single person are denoted $\mathbf{x} = \{\phi, \theta, \beta, \tau\}$.
The reconstructed results must primarily adhere to visual evidence. For the visible parts, we leverage the existing HMR framework~\cite{ge2021yolox, AutoTrackAnything, goel2023humans4D} to obtain initial estimates and image features. We first employ off-the-shelf human detection and tracking methods~\cite{AutoTrackAnything, cheng2022xmem, mobile_sam, yolov8_ultralytics, ge2021yolox} to acquire the bounding boxes of individuals in images, and then a Vision Transformer (ViT)~\cite{goel2023humans4D} pretrained on extensive datasets serves as the backbone network to extract image features $F_{img}$ within these bounding boxes. 
We further apply a motion head with sequential MLP layers to obtain SMPL tokens from $F_{img}$, and derive the initial coarse estimates $\mathbf{x} = \{x_a, x_b\}$. For the interactive descriptions generated by the VLM Annotator, we use CLIP~\cite{radford2021clip} as the text encoder to obtain textual features $F_{\text{text}}$. We incorporate semantic cues to facilitate complete reconstruction of occluded body regions.

\noindent \textbf{Interactive Diffusor.}
The Interactive Diffusor integrates visual features from observable body regions with textual semantic guidance to generate interactive motions. Recent advancements in controllable diffusion-based generation~\cite{xie2023omnicontrol,wan2024tlcontrol,zhang2023adding} are ideal for our task: visible body regions require strict adherence to input images, while occluded regions need context-aware completion. Unlike traditional diffusion models that start from pure noise, we generate from the coarse motion $\mathbf{x}$ with distribution adaptation~\cite{huang2024closeInt}, which ensures output consistency with observed human poses and preserves key pose features in the generated results.

Following prior methods~\cite{wang2023intercontrol, xie2023omnicontrol}, the Interactive Diffusor takes the interactive individual motions $x_a^t$ and $x_b^t$ extracted from SMPL head as denoising inputs. It then produces the corresponding denoised motions $\hat{x}_a^0$ and $\hat{x}_b^0$ conditioned on the diffusion timestep $t$ and image features $F_{img}$. The textual descriptions serve as auxiliary guidance through a zero-initialized layer, similar to ControlNet~\cite{zhang2023adding}. Human interactions inherently involve mutual influence between individuals' movements. To model this, we adopt a dual-branch structure with cross-attention mechanisms~\cite{liang2024intergen}, where each branch handles motion reconstruction for one individual while maintaining shared weights and bidirectional information exchange. This configuration effectively captures the reciprocal nature of interactive motions.
Details are provided in the Appendix.

\noindent \textbf{Model Training.}
We optimize via the following objective function:
$
\mathcal{L}=\mathcal{L}_{\mathrm{reproj}}+\mathcal{L}_{\mathrm{smpl}}+\mathcal{L}_{\mathrm{joint}}+\mathcal{L}_{\mathrm{vel}}+\mathcal{L}_{\mathrm{int}}+\mathcal{L}_{\mathrm{pen}}
$
, where $\mathcal{L}_{\mathrm{reproj}}=\|\Pi\left (J+\tau\right)-\hat{J_{2D}}\|_2^2$ measures the discrepancy between projected 3D joints and 2D ground-truth poses; $J \in \mathbb{R}^{21 \times 3}$ denotes 3D joint positions derived from SMPL parameters. $\mathcal{L}_{\mathrm{smpl}}$, $\mathcal{L}_{\mathrm{joint}}$, and $\mathcal{L}_{\mathrm{vel}}$ are $\mathcal{L}_{2}$ distances between predicted and target shape parameters, 3D joint positions, and joint velocities, respectively. 
$
    \mathcal{L}_{\mathrm{int}}=\left\|\left|J_a-J_b\right|-\left|\hat{J}_a-\hat{J}_b\right|\right\|_2^2$
supervises inter-person joint distances. For the penetration loss, we first detect the set of colliding triangles using bounding volume hierarchies (BVH)~\cite{karras2012maximizing}, then compute $\mathcal{L}_{\mathrm{pen}}$ as:
\begin{equation}
    \tiny
    \mathcal{L}_{\mathrm{pen}}=\sum_{ (f_a,f_b)\in\mathcal{C}}\left\{\sum_{v_a \in f_a}\left\|-\Psi_{f_b}\left (v_a\right)n_a\right\|^2+\sum_{\upsilon_b \in f_b}\left\|-\Psi_{f_a}\left (v_b\right)n_b\right\|^2\right\}
\end{equation}
, where $f_a$ and $f_b$ are two colliding triangles in the set $\mathcal{C}$. Here $v$ and $n$ denote vertex positions and normals, respectively, and $\Psi (\cdot)$ is the distance field.

\subsection{Geometry Optimizer}
\label{sec:geo}

In the diffusion stage, each frame is represented with SMPL parameters and camera-space root motion. This parameterization can under-specify fine-grained spatial relations between subjects, whereas explicit 3D joint trajectories make relative layout and contact easier to supervise. We therefore attach an auxiliary module that regresses the 3D joints positions of both people along the sequence and uses them to refine the final motion estimate.

\noindent \textbf{Auxiliary Module.}
The Auxiliary Module adopts the same two-branch mutual attention structure as the diffusion model. The key difference is that the linear layer in the Motion Embedding component is replaced with a Spatial-Temporal Graph Convolutional Network (STGCN)~\cite{yu2017spatio, Yan2018SpatialTG}, which models spatio-temporal relationships. Based on the human anatomical structure, nodes in each frame are connected to form spatial edges, while temporal edges link corresponding joints across consecutive time steps. This setup enables the construction of multi-layer spatial-temporal graph convolutions, facilitating the integration of information across both spatial and temporal dimensions. We convert the joint pair annotations from the VLM Annotator into a contact mask $\mathrm{M} \in \mathcal{R}^{\mathrm{K} \times \mathrm{L}}$, with $M_{k,l} = 1$ if there is contact and $M_{k,l} = 0$ if there is no contact. This model is trained with a composite loss function defined as $\mathcal{L} = \mathcal{L}_{\text{reproj}} + \mathcal{L}_{contact} + \mathcal{L}_{\text{vel}} + \mathcal{L}_{\text{int}}$, where $\mathcal{L}_{contact} = (\alpha \mathrm{M} + \mathbf{1}_{K \times L})\mathcal{L}_{\text{joint}}$, which deliberately strengthens the contact positions and 3D geometric information, thereby enhancing the spatial relationships.

\subsection{Temporal Motion Refiner}
\label{sec:tem}
Through the Semantic-Guided Motion Infiller, we generate an interaction motion sequence conditioned on both visual and semantic cues. However, textual guidance may unintentionally alter visible regions, and interpenetration artifacts can still occur due to the use of soft collision penalties. To address this issue, we apply a confidence-based infilling strategy. 
Given the initial estimates $\mathbf{x}$ from SMPL head, the infilled sequence $\hat{x}^0$, and a confidence mask $\mathcal{C}$, the final motion sequence is obtained as
$x' \;=\; M\odot\mathbf{x} \;+\; (1 - M)\odot \hat{x}^0$,
where $M = \mathbf{1}_{\{\mathcal{C} \geq \theta\}}$ is a binary mask derived from the confidence scores $\mathcal{C} \in [0,1]^T$ with threshold $\theta \in [0,1]$, and $\odot$ denotes element-wise multiplication.
Thus, through this operation, diffusion infilling mainly affects low-confidence regions, while high-confidence regions are preserved from the initial estimates.
Subsequently, we further optimize the infilled motion sequence with a frozen Interactive Diffusor, which leverages the generative prior of diffusion models to naturally improve temporal coherence and produce smoother transitions across frames. In parallel, it integrates guidance from the Geometry Optimizer to refine spatial relationships, leading to motion sequences that are both geometrically consistent and temporally smooth.
Additionally, we introduce a factorized loss guidance that enables joint constraints and collision-guided sampling, and independently optimizes body shape and joints for efficient convergence.

\noindent \textbf{Factorized Loss Guidance.} 
During the sampling process, we incorporate joint and collision guidance signals to improve interaction quality. The joint guidance signal is derived from the joint positions generated by the Geometry Optimizer. For the predicted motion $\hat{\mathbf{x}}_0$ from the frozen Interactive Diffusor, we derive its 3D joint positions $J$ and compute the weighted contact loss between $J$ and the guidance signal $J'$ from the Geometry Optimizer as $\mathcal{L}_{\text{contact}} (J,J') = (\alpha \mathrm{M} + \mathbf{1}_{K \times L})||J-J'||^2$.
The collision guidance signal is based on the interpenetration volume between meshes, which reduces mesh penetration, improving the geometric plausibility of the interaction. We reconstruct meshes for the two individuals from $\hat{\mathbf{x}}_0$ and use BVH to compute their $L_2$ intersection volume difference as $\mathcal{L}_{\text{penetration}}$.
The guidance loss is defined as $\mathcal{L}_{\text{guidance}} = \lambda_j \mathcal{L}_{\text{contact}} + \lambda_p \mathcal{L}_{\text{penetration}}$, where $\lambda_j$ and $\lambda_p$ are weighting parameters.
Following the methodology in InterControl~\cite{wang2023intercontrol}, we perform multiple L-BFGS iterations at each denoising step to update the posterior mean. The optimization process is described as follows:
$\mu_t' = \mu_t - \lambda \nabla_{\mu_t}\mathcal{L}_{\text{guidance}} (\mu_t)$,
where $\lambda$ denotes the optimization step size.

In addition, joint optimization of heterogeneous parameters (rotation, shape, and translation) with uniform settings leads to suboptimal outcomes. To address this, we introduce a factorized loss guidance approach. Since joint guidance provides limited shape-related information and collision constraints may cause unwanted morphological compression when applied to shape parameters, we decompose the optimization process into two components: rotational parameters $\mu_t^{\text{pose}}$ and translational parameters $\mu_t^{\text{transl}}$, for separate optimization.
Each component undergoes multi-round iterative optimization with the L-BFGS optimizer.
This factorized approach allows for task-specific optimization, leading to more efficient convergence and more plausible results.

\begin{figure*}[t]
    \centering
    \includegraphics[width=\linewidth]{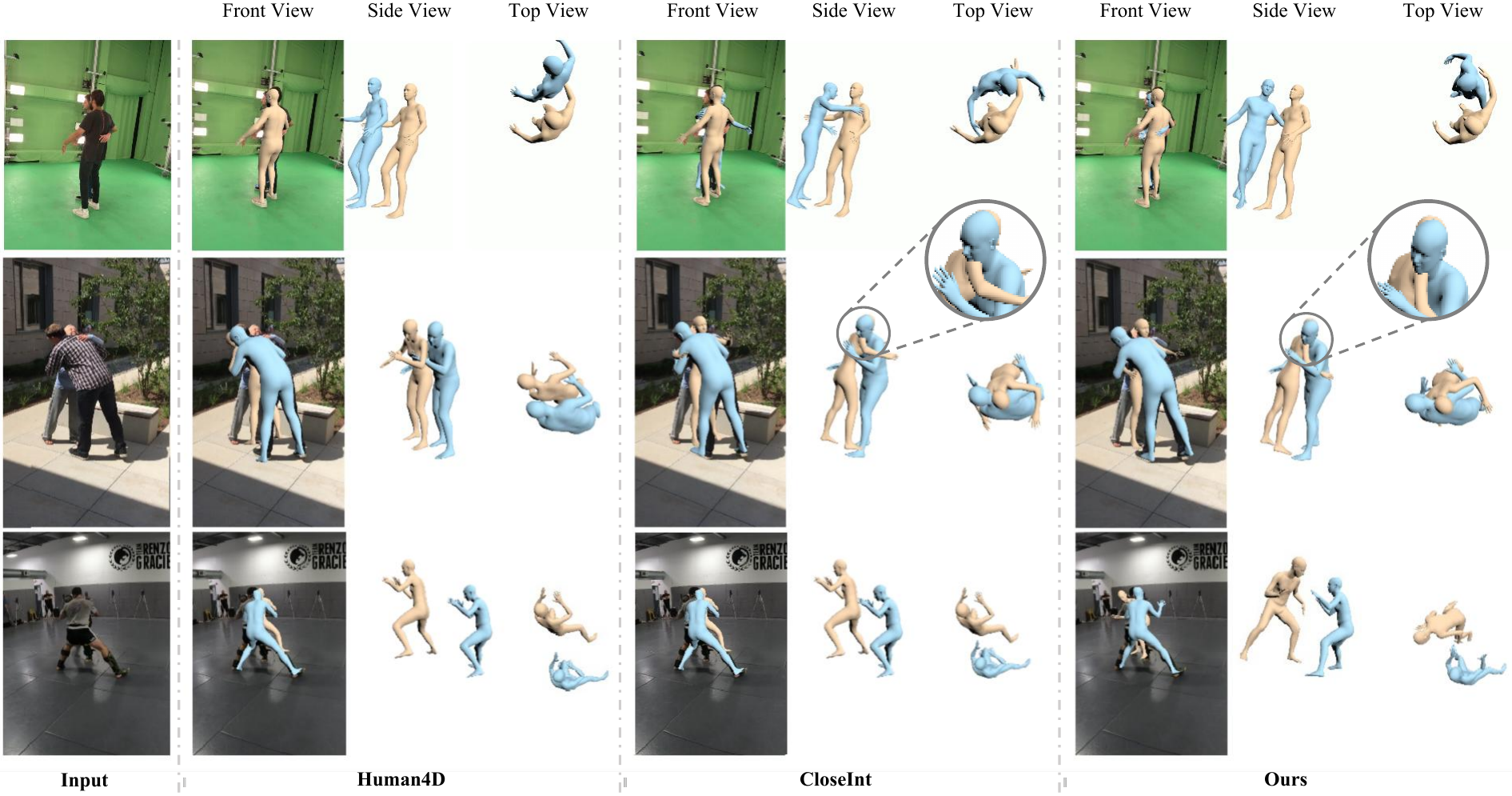}
    \vspace{-4ex}    
    \caption{Qualitative comparison results.
    }
    \label{fig:vis-result}
    \vspace{-2ex}
\end{figure*}

\section{Experiments}

\subsection{Datasets}

\noindent \textbf{Hi4D}~\cite{yin2023hi4d} focuses on close human interaction scenarios, encompassing dynamic interaction types such as hugging, dancing, and athletic movements. It challenges existing methods' capacity to handle prolonged occlusions and complex interactions. The dataset comprises 20 unique participant pairs, totaling 100 sequences with over 11,000 frames, of which more than 6,000 frames contain physical contact. For consistency, we adopt the same training and test split protocol as the baseline.

\noindent \textbf{3DPW}~\cite{von2018recovering} records human activities in natural environments, encompassing various daily scenarios such as courtyards, downtown areas, and offices. We selected sequences involving two-person interactions from these recordings, resulting in a total of 31 sequences with 12,000 frames.

\noindent \textbf{Harmony4D}~\cite{khirodkar2024harmony4d} is a multi-view video dataset specialized in in-the-wild close human interactions. Unlike datasets collected in controlled settings with choreographed motions, Harmony4D captures naturally occurring dynamic activities including wrestling, dancing, and mixed martial arts. The dataset contains 208 video sequences captured by over 20 synchronized cameras, yielding 1.66 million images across five distinct scenarios involving 24 unique participants. 
We utilize the test set of this dataset to validate the generalization ability of our model on unseen datasets without training.

\begin{table*}[t]
\vspace{-4ex}
\centering
\caption{Comparisons on Hi4D and 3DPW. PA denotes PA-MPJPE and VPE denotes MPVPE.}
\vspace{-2ex}
\label{tab:Comparisons on Hi4D and 3DPW.}
\tiny
\resizebox{\linewidth}{!}{
    \begin{tabular}{l|cccccccc|cccccccc} 
    \toprule
    \multirow{2}{*}{Method} & \multicolumn{8}{c|}{Hi4D} & \multicolumn{8}{c}{3DPW} \\
    \cmidrule (lr){2-9} \cmidrule (l){10-17}
    & $\downarrow$RE & $\downarrow$GE & $\downarrow$Int & $\downarrow$Smoothness & $\downarrow$Pen &  $\downarrow$MPJPE & $\downarrow$PA & $\downarrow$VPE & $\downarrow$RE & $\downarrow$GE & $\downarrow$Int & $\downarrow$Smoothness & $\downarrow$Pen & $\downarrow$MPJPE & $\downarrow$PA & $\downarrow$VPE \\
    \midrule
    Human4D~\cite{goel2023humans4D} & - & - & - & -  & - & 72.1 & 52.4 & 88.6 & - & - & - & - & - & 72.9 & 49.1 & 107.0\\
    BEV~\cite{sun2022BEV} & 210.5 & 223.5 & 131.0 & - & 1953.6  & 91.8 & 59.3 & 101.2 & 235.2 & 291.8 & 145.6 & - & 233.8  & 135.0 & 81.9 & 169.7\\
    GroupRec~\cite{Huang2023ReconstructingGO} & 113.2 & 122.3 & 98.8 & - & \textbf{1858.4}  & 82.4 & 51.6 & 88.6 & 204.6 & 235.2 & 110.6 & - & \textbf{100.9}  & 73.3 & 48.7 & 109.4\\
    BUDDI~\cite{muller2024buddi}  & 200.3 & 216.4 & 102.6 & - & 1879.3  & 96.8 & 70.6 & 116.0 & 228.4 & 289.4 & 113.1 & - & 203.5 & 83.6 & 53.6 & 93.8 \\
    CloseInt~\cite{huang2024closeInt} & 99.0  & 114.9  & 81.4 & 4.6 & 3947.6   & 63.1  & \textbf{47.5}  & \textbf{76.4}  & 121.1 & 134.0 & 75.6 & 19.9 & 101.6  & 59.0 & 45.3 & 73.2\\
    Ours & \textbf{83.6}  & \textbf{95.2}  & \textbf{68.5} & \textbf{3.5} & 2380.5 & \textbf{62.2}  & \textbf{47.5}  & 79.3  & \textbf{91.0} & \textbf{127.9} & \textbf{64.6} & \textbf{10.0} & 109.7  & \textbf{55.6} & \textbf{44.3} & \textbf{69.4}\\
    \bottomrule
    \end{tabular}
}
\end{table*}

\begin{table*}[ht]
  \vspace{-5ex}
  \noindent
    \begin{minipage}{0.5\linewidth}
    \caption{Comparisons on Harmony4D.}
    \vspace{-2ex}
    \label{tab:Comparisons on Harmony4D.}
    \tiny
    \resizebox{\linewidth}{!}{
      \begin{tabular}{l|cccccccc} 
        \toprule
        & $\downarrow$RE & $\downarrow$GE & $\downarrow$Int & $\downarrow$Pen & $\downarrow$MPJPE & $\downarrow$PA & $\downarrow$VPE \\
        \midrule
        Human4D & - & - & - & -  & 108.2 & 60.3 & 131.0 \\
        BEV & 365.4 & 716.7 & 360.4  & 484.4 & 111.3 & 78.0 & 144.3 \\
        GroupRec & 346.6 & 689.2 & 337.1 & 499.4 & 119.0 & 65.5 & 144.8 \\
        BUDDI &352.3 & 692.3 & 324.1 & \textbf{479.3} & 126.4 & 84.0 & 158.7 \\
        CloseInt & 202.2 & 446.6 & 255.2 & 488.9 & \textbf{103.5} & 47.1 & \textbf{114.9} \\
        Ours & \textbf{198.2} & \textbf{411.8} & \textbf{245.6} & 482.9 & 104.6 & \textbf{45.9} & 117.3 \\
        \bottomrule
      \end{tabular}
    }
    \vspace{-2ex}
    \end{minipage}
    \hfill  
    \begin{minipage}{0.4\linewidth}
      \centering
      \includegraphics[width=\linewidth]{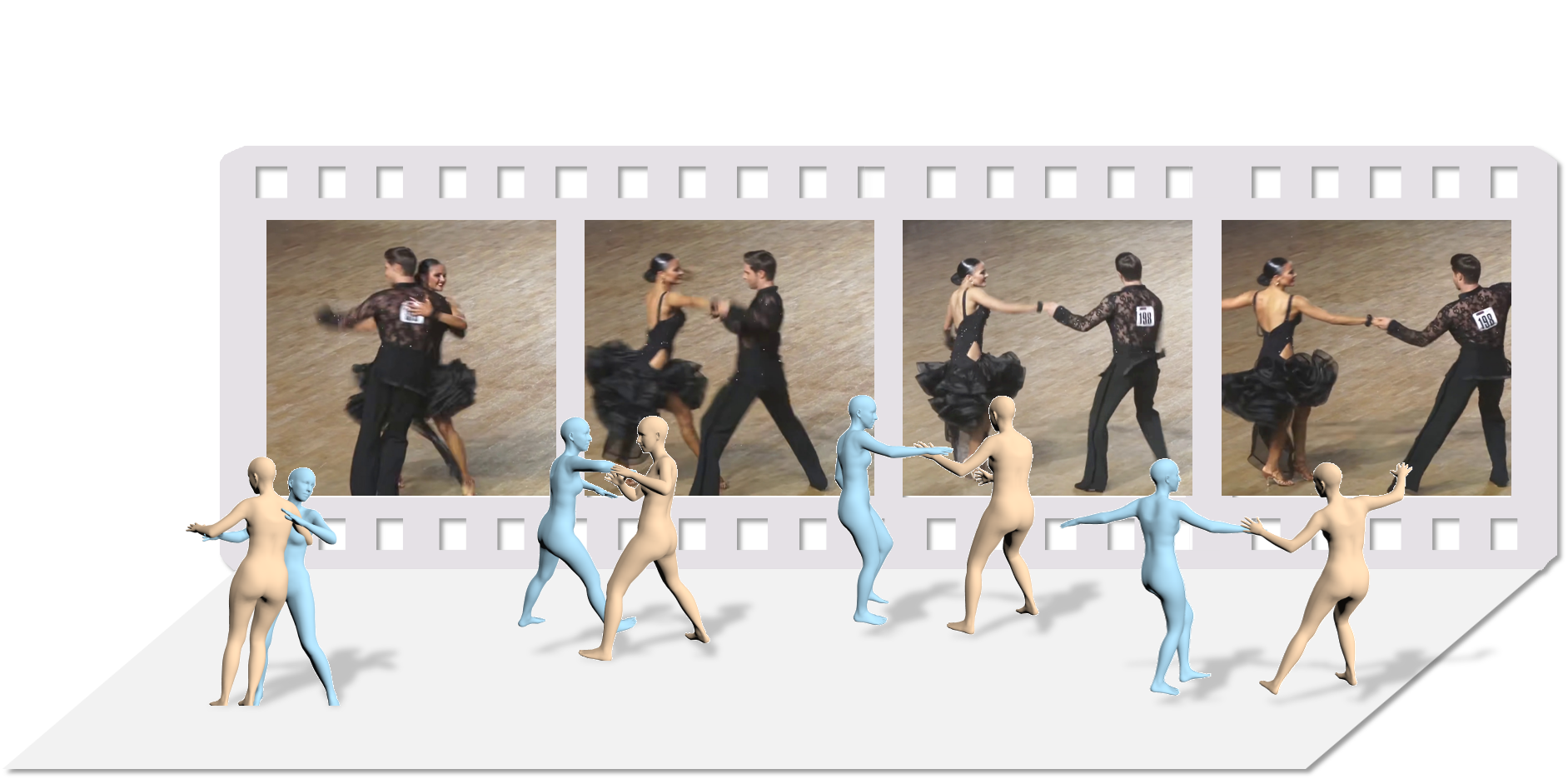} 
      \vspace{-4ex}
      \captionof{figure}{Visualization on in-the-wild video.}
      \label{fig:exp-wild}
    \vspace{-2ex}
    \end{minipage}
\end{table*}

\begin{table*}[t]
\caption{Ablation studies on the impact of semantic and geometric information on Hi4D. Temporal Motion Refiner with a dash (-) indicates Temporal Motion Refiner without factorized loss guidance and contact mask.
}
\vspace{-2ex}
\centering
\tiny
\resizebox{\linewidth}{!}{
\begin{tabular}{ccccc|cccccccccccc}
\toprule
\multicolumn{1}{c|}{Semantic-Guided}  & \multicolumn{3}{c|}{Temporal Motion Refiner} & \multicolumn{1}{c|}{Geometry} & \multirow{2}{*}{$\downarrow$RE} & \multirow{2}{*}{$\downarrow$GE} & \multirow{2}{*}{$\downarrow$Int} & 
 \multirow{2}{*}{$\downarrow$Smoothness} & \multirow{2}{*}{$\downarrow$MPJPE} & \multirow{2}{*}{$\downarrow$PA}  & \multirow{2}{*}{$\downarrow$VPE} \\
\multicolumn{1}{c|}{Motion Infiller} & \multicolumn{1}{c|}{-}  & \multicolumn{1}{c|}{factorized loss guidance} & \multicolumn{1}{c|}{contact mask} & \multicolumn{1}{c|}{Optimizer} \\
\midrule
  & & & & & 100.4 & 119.0 & 90.5 & 4.7 & 62.4 & 47.5 & \textbf{78.3} \\
\checkmark & & & & & 91.2  & 102.7 & 73.4 &  4.1 &  63.5 & 48.7 & 80.7 \\
\checkmark & \checkmark & & & & 91.5 & 103.5 & 74.2 & 3.9 & 63.3 & 48.5 & 80.2 \\
\checkmark & \checkmark & & & \checkmark & 91.0  & 102.4 & 73.3 &  4.1 & 63.5 & 48.7 & 80.7 \\
 & \checkmark & \checkmark & & \checkmark & 89.3 & 100.6 & 69.2 & 4.0 & 63.2 & 48.5 & 80.3 \\
 & \checkmark & \checkmark & \checkmark & \checkmark & 88.6 & 98.7 & 68.7 & 3.9 & 63.0 & 48.5 & 80.2 \\
 \checkmark & \checkmark & \checkmark & & \checkmark & 84.5  & 96.2  & \textbf{68.5} & \textbf{3.5} & 62.8 & 47.7 & 79.8 \\
 \textbf{\checkmark} & \textbf{\checkmark} & \textbf{\checkmark} & \textbf{\checkmark} & \textbf{\checkmark} & \textbf{83.6} & \textbf{95.2} & \textbf{68.5} & \textbf{3.5} & \textbf{62.2} & \textbf{47.5} & 79.3 \\
\bottomrule
\end{tabular}}
\label{tab:ablation}
\vspace{-2ex}  
\end{table*}

\definecolor{blue1}{rgb}{0.97647,0.992156,0.99607}
\definecolor{blue2}{rgb}{0.84313,0.945098, 0.968627}
\definecolor{blue3}{rgb}{0.713725,0.89803,0.94509}
\definecolor{blue4}{rgb}{0.58039,0.85098,0.921568}
\definecolor{blue5}{rgb}{0.44705,0.80784,0.89803}

\begin{table*}[t]
    \centering
    \caption{Results under various occlusion severities on Hi4D. Darker colors indicate greater improvements. Improve. is improvement rate.}
    \vspace{-2ex}
    \label{tab:Results under various occlusion severities}
    \tiny
    \resizebox{\linewidth}{!}{
     \begin{tabular}{l|ccc|ccc|ccc|ccc|ccc} 
    \toprule
    \multirow{1}{*}{IoU} & \multicolumn{3}{c|}{0.0} & \multicolumn{3}{c|}{$ (0,0.25]$} & \multicolumn{3}{c|}{$ (0.25,0.5]$} & \multicolumn{3}{c|}{$ (0.5,0.75]$} & \multicolumn{3}{c}{$ (0.75,1.0]$} \\
    \cmidrule (lr){2 - 4} \cmidrule (lr){5 - 7} \cmidrule (lr){8 - 10} \cmidrule (lr){11 - 13} \cmidrule (l){14 - 16}
    & CloseInt & Ours & Improve. & CloseInt & Ours & Improve. & CloseInt & Ours & Improve. & CloseInt & Ours & Improve. & CloseInt & Ours & Improve. \\
    \midrule
    $\downarrow$RE & \cellcolor{blue1}85.4 & \cellcolor{blue1}78.3 & \cellcolor{blue1}8.3 & \cellcolor{blue3}93.3 & \cellcolor{blue3}82.3 & \cellcolor{blue3}11.8 & \cellcolor{blue5}104.1 & \cellcolor{blue5}91.4 & \cellcolor{blue5}12.2 & \cellcolor{blue4}104.3 & \cellcolor{blue4}93.8 & \cellcolor{blue4}10.1 & \cellcolor{blue2}108.4 & \cellcolor{blue2}106.7 & \cellcolor{blue2}1.6 \\
    
    $\downarrow$GE & \cellcolor{blue2}99.4 & \cellcolor{blue2}89.0 & \cellcolor{blue2}10.5 & \cellcolor{blue4}112.4 & \cellcolor{blue4}96.3 & \cellcolor{blue4}14.3 & \cellcolor{blue5}120.4 & \cellcolor{blue5}101.2 & \cellcolor{blue5}16.0 & \cellcolor{blue3}118.3 & \cellcolor{blue3}105.3 & \cellcolor{blue3}11.0 & \cellcolor{blue1}125.6 & \cellcolor{blue1}117.7 & \cellcolor{blue1}6.3 \\
    
    $\downarrow$Int & \cellcolor{blue1}67.8 & \cellcolor{blue1}62.6 & \cellcolor{blue1}7.7 & \cellcolor{blue5}87.2 & \cellcolor{blue5}68.6 & \cellcolor{blue5}21.3 & \cellcolor{blue4}98.2 & \cellcolor{blue4}79.3 & \cellcolor{blue4}19.3 & \cellcolor{blue3}93.3 & \cellcolor{blue3}77.5 & \cellcolor{blue3}16.9 & \cellcolor{blue2}93.3 & \cellcolor{blue2}82.4 & \cellcolor{blue2}11.7 \\
    
    
    $\downarrow$MPJPE & \cellcolor{blue1}40.9 & \cellcolor{blue1}44.7 & \cellcolor{blue1}-9.3 & \cellcolor{blue3}49.2 & \cellcolor{blue3}51.3 & \cellcolor{blue3}-4.3 & \cellcolor{blue4}64.2 & \cellcolor{blue4}65.7 & \cellcolor{blue4}-2.3 & \cellcolor{blue5}74.6 & \cellcolor{blue5}72.6 & \cellcolor{blue5}2.7 & \cellcolor{blue2}84.1 & \cellcolor{blue2}88.1 & \cellcolor{blue2}-4.8 \\
    
    $\downarrow$PA & \cellcolor{blue1}31.1 & \cellcolor{blue1}35.2 & \cellcolor{blue1}-13.2 & \cellcolor{blue2}37.6 & \cellcolor{blue2}40.1 & \cellcolor{blue2}-6.7 & \cellcolor{blue3}49.3 & \cellcolor{blue3}51.3 & \cellcolor{blue3}-4.1 & \cellcolor{blue5}55.6 & \cellcolor{blue5}54.9 & \cellcolor{blue5}1.3 & \cellcolor{blue4}58.6 & \cellcolor{blue4}60.6 & \cellcolor{blue4}-3.4 \\
    
    \bottomrule
    \end{tabular}}
    \vspace{-4ex}
\end{table*}

\subsection{Evaluation Metrics}

We adopt the metrics of CloseInt~\cite{huang2024closeInt}, including Root-Aligned Mean Per Joint Position Error (\textbf{MPJPE}) and Procrustes-Aligned MPJPE (\textbf{PA-MPJPE}) for per-person pose accuracy, and Mean Per Vertex Position Error (\textbf{MPVPE}) for mesh error. Together they summarize single-subject reconstruction quality.
To further evaluate the network's capacity to model spatial relationships in multi-person interactions, in addition to the \textbf{Interaction} metric defined in CloseInt, we introduce two complementary metrics: Global Mean Per Joint Position Error (\textbf{G-MPJPE (GE)}) measuring absolute pose errors across the entire scene, and Relative Mean Per Joint Position Error (\textbf{R-MPJPE (RE)}) focusing on inter-person positional relationships. RE is defined as the mean per joint position error after aligning to the first person's root position. It removes global translation between subjects and therefore emphasizes relative joint placement.
For temporal consistency, we follow MultiPhys~\cite{ugrinovic2024multiphys} and report \textbf{Smoothness}, calculated as the mean squared error between the predicted and ground-truth accelerations of each joint. This metric quantifies the continuity of joint movements across the temporal sequence.
For physical plausibility, we report inter-person penetration volume (\textbf{Pen}), quantified by computing the signed distance function (SDF) for each subject and accumulating the penetration depth across intersecting vertices. The Pen metric represents the average sum of negative SDF values per person over the entire sequence.

\subsection{Main Results}

\noindent \textbf{Results on Hi4D and 3DPW.} We compare our method with several state-of-the-art baseline methods on Hi4D~\cite{yin2023hi4d} and 3DPW~\cite{von2018recovering}. A dash (-) indicates that some results are either not reported or unavailable. 

While Human4D~\cite{goel2023humans4D} achieves promising results on single-person metrics, it does not account for mutual relationships of interacting individuals and fails to capture spatial dependencies between different subjects. BEV~\cite{sun2022BEV} and GroupRec~\cite{Huang2023ReconstructingGO} explicitly consider depth relationships among people to mitigate depth ambiguity in monocular multi-person reconstruction, yet they struggle with complex interaction patterns in close-contact settings. BUDDI~\cite{muller2024buddi} and CloseInt~\cite{huang2024closeInt} share the most similar settings with our method, focusing on monocular two-person reconstruction under close interaction. 
BUDDI uses a Generative Proxemics model to align meshes with the initial estimate and detected keypoints. The quality of its results relies on the accuracy of the keypoints, which can be unreliable or missing when humans are heavily occluded. Additionally, BUDDI lacks temporal modeling for handling dynamic interactions over time.
CloseInt employs a two-person interaction prior, which also relies on the precondition that the movements of both individuals can be roughly reconstructed from images. Therefore, it struggles with heavy occlusions in monocular videos, resulting in poor interaction modeling.

In contrast to these methods, our approach introduces semantic information to infill occluded body parts, operating without reliance on 2D keypoint detection or flawed image features. 
We achieve 4.2\% and 18.3\% relative improvements in RE and Int compared with the latest SOTA on Hi4D. It is critical to clarify the interpretation of single-person vs. interaction-focused metrics here: MPJPE, PA, and VPE focus solely on per-person reconstruction accuracy. Due to root alignment in their computation, they cannot capture errors in positioning or root jitter, which are critical for evaluating interaction quality. This inherent limitation explains why our method shows only marginal changes in MPJPE and VPE. By contrast, the substantial gains in RE, GE, and Int directly validate that SocialMirror effectively addresses the challenges of severe mutual occlusions and disrupted spatial relationships, which are the primary pain points of monocular interaction reconstruction.

Inter-person penetration should not be analyzed in isolation, incorrectly placing two people far apart in a close-interaction clip can also yield a small Pen value. Moreover, slight mesh penetration is often acceptable in tight contact (e.g., hugging), where, for instance, one person's palm may slightly intersect the other person's body. To assess whether a method reduces penetration while correctly inferring relative placement, RE, Int, and Pen should be considered jointly.
Our method simultaneously achieves low RE, Int, and Pen, indicating that it captures relative spatial structure while keeping penetration moderate.

\noindent \textbf{Generalization Evaluation.} 
We further assess SocialMirror's generalization on the unseen Harmony4D dataset~\cite{khirodkar2024harmony4d} without fine-tuning (Table~\ref{tab:Comparisons on Harmony4D.}). SocialMirror improves RE, GE, Int, and Pen over most baselines while remaining close to the best single-person errors, indicating robust interaction modeling under domain shift. 
Qualitatively, Figs.~\ref{fig:teasor} and~\ref{fig:exp-wild} show reconstructions on in-the-wild human interaction clips collected from the web. Under heavy occlusion, our estimates still exhibit plausible articulation and mutual contact. Further examples are provided in the Appendix; together with the Harmony4D evaluation, they suggest that SocialMirror transfers to web video outside the curated training domains.
\noindent \textbf{VLM Annotation User Study.} 
Semantic information plays a vital role in our framework by complementing motion reconstruction. To assess the quality of interaction descriptions generated by the VLM Annotator, we conducted a user study with 20 participants on 40 randomly selected video sequences. Participants rated alignment between the VLM-generated text and the video on a 5-point scale (1: irrelevant or incorrect; 3: comparable to typical human annotations in correctness; 5: exceptionally accurate). The VLM Annotator achieved a mean score of 3.3, indicating reliable annotations. In addition, participants consistently reported that VLM descriptions were more detailed than human, often spelling out body-part mentions and fine-grained motion phases that human annotators summarized more briefly, which is beneficial for semantic guidance in our pipeline.

\subsection{Ablation Study}

\noindent \textbf{Module ablations.}
We conducted ablation studies to evaluate the impact of different modules in Table~\ref{tab:ablation}.
Introducing the Semantic-Guided Motion Infiller module leads to a notable performance improvement, particularly in reducing GE and RE. 
The semantic information integration enables the network to preserve critical visual features while incorporating textual descriptions, leading to a more accurate recovery of interactive motions and spatial relationships.

Introducing the Temporal Motion Refiner without factorized loss guidance or contact mask improves motion smoothness but degrades RE, GE, and Int, confirming that temporal smoothing alone fails to resolve spatial inaccuracies. 
The Geometry Optimizer alone yields only modest improvements. In contrast, consistent improvements emerge when factorized loss guidance is employed: decoupling rotation and translation allows independent tuning of hyperparameters for each parameter and thus leads to superior convergence and overall reconstruction quality.
Adding a contact mask further reduces RE, GE, and Int It primarily refines local details, such as hand–contact interactions. These fine-grained adjustments, which are typically centimeter-scale in localized regions, often manifest as subtle metric improvements that may not appear pronounced numerically.

When all modules are combined, the full model achieves optimal performance, with the most significant gains observed in interaction-related metrics and motion smoothness. This validates the synergistic effect of each module.
Notably, MPJPE, PA, and VPE metrics remain relatively stable across configurations, suggesting the model prioritizes global motion realism over joint-level precision, which is a trade-off favorable for visually realistic reconstructions. 

\noindent \textbf{Comparison of various occlusion severities.} 
To further explore the effectiveness of our method on different occlusion levels, we quantified occlusion severity by computing intersection-over-union (IoU) between bounding boxes, partitioning the test set into five subsets representing distinct occlusion levels. As shown in Table~\ref{tab:Results under various occlusion severities}, our approach achieves comparable results to CloseInt in scenarios without occlusion and consistently outperforms it under partial and moderate occlusion (IoU between 0.25 and 0.75). 
With semantic and geometric guidance, the model reconstructs plausible poses by leveraging VLM-generated textual descriptions when visual cues are lacking, while also improves the interaction quality and spatial accuracy.
Therefore, we obtain more natural and realistic real-world human interactions under challenging occlusion conditions.

\section{Conclusion}

We present SocialMirror, a diffusion-based method that integrates semantic cues and geometric constraints to address the challenges of monocular human mesh reconstruction in close interaction scenarios. 
The Semantic-Guided Motion Infiller leverages vision-language descriptions to reconstruct occluded regions and resolve pose ambiguities. The Geometry Optimizer and the Temporal Motion Refiner enforce 3D joint consistency and temporal consistency, enhancing spatial plausibility and natural contact relationships.
Extensive evaluations demonstrate that SocialMirror delivers realistic, semantically enriched reconstructions across various datasets and in-the-wild scenarios.


{
    \small
    \bibliographystyle{ieeenat_fullname}
    \bibliography{sec/citation}
}

\clearpage
\setcounter{page}{1}
\maketitlesupplementary

We present additional implementation details, including model setup, dataset processing, diffusion process modification, and the two-branch network architecture, as well as VLM annotator details with prompting examples in Sec.~\ref{sec:add_det}. Additional experiments are provided in Sec.~\ref{sec:add_exp}, including an ablation on the motion embedding layer in the Geometry Optimizer, performance breakdowns across Hi4D's action categories,
cross-dataset results, and in-the-wild visualizations. Sec.~\ref{sec:discussions} includes analyses of VLM limitations and failure cases, while also exploring the role of semantic information in limited-contact scenarios and outlining the framework's current limitations. The use of Large Language Models are declared in Sec.\ref{sec:use_llm}

\section{Additional Details}
\label{sec:add_det}

\subsection{Implementation details}

Our model was implemented using PyTorch and trained on an NVIDIA RTX 3090 GPU. The batch size was set to 32 for the Semantic-Guided Motion Infiller and 64 for the Geometry Optimizer. We employed the AdamW optimizer with CyclicLRWithRestarts, where the learning rate was initially set to 0.0001, with parameters restart\_period=10, t\_mult=2, and a "cosine" policy.

In the Motion Infiller and Motion Refiner, the dimension of human motion followed CloseInt~\cite{huang2024closeInt} with D = 157. For the Geometry Optimizer, we utilized 24 SMPL joints to represent human motion, resulting in a human motion dimension of D' = 24 × 3. The text feature dimension $F_{text}$, encoded from CLIP~\cite{radford2021clip}, was 256.

For dataset implementation, original long motion sequences were divided into shorter clips with a length of L = 16 frames. Each clip was annotated with a corresponding text description using our LLM annotation module. For 3DPW, we established a new benchmark by selecting sequences involving two subjects: sequences captured in courtyard environments were used for training, and those captured in downtown settings were used for testing.

For multi-person scenes, we automatically detect and track individuals to obtain their bounding boxes and select the pair with the closest spatial proximity as the primary subjects. The original image is then cropped according to their bounding boxes, centering the region of interest to minimize background distractions and ensure the VLM focuses exclusively on the targets.

\subsection{Diffusion with initial distributions}

In prior approaches to diffusion-based pose estimation~\cite{feng2023diffpose, rommel2023diffhpe}, time-dependent Gaussian noise sampled from $\mathcal{N}\left (0,\mathrm{I}\right)$ is incrementally injected into ground-truth motion sequences $\hat{x}_0$ through the forward process:
\begin{equation}
q (\mathbf{x}_t\mid\hat{\mathbf{x}}_0)=\sqrt{\hat{\alpha}_t}\hat{\mathbf{x}}_0+\sqrt{1-\hat{\alpha}_t}\epsilon,\epsilon\sim\mathcal{N} (0,1)
\end{equation}

where $\alpha_t$ denotes a constant hyper-parameter~\cite{nichol2021improved}, and $\hat{\alpha}_t = \prod_{i=0}^{t} \alpha_i$. It was observed that $x_t$ follows a standard Gaussian distribution, and the early iterative steps provide limited meaningful information for human motion. Additionally, the results should fully account for the initial prediction consistent with image characteristics. 

To address these issues, we propose modifying the forward diffusion process to align with the initial distributions:

\begin{equation}
\footnotesize
    q (x_t | \hat{x}_0) = x + \sqrt{\hat{\alpha}_t} (\hat{x}_0 - x) + \sqrt{1 - \hat{\alpha}_t} \epsilon, \quad \epsilon \sim \mathcal{N} (0, \sigma)
\end{equation}

With this adjusted framework, a generative model is derived by reversing the diffusion process, starting from samples $x_t \sim \mathcal{N} (x, \sigma)$. The reverse process is defined as:

\begin{equation}
    q (x_{t-1} | x_t, c) = \mathcal{N} \left ( x_{t-1}; \mu_{\alpha} (x_t, c), \tilde{\beta}_t \sigma \right)
\end{equation}

where $\mu_{\alpha} (x_t, c)$ represents the estimated mean from the diffusion model under condition $c$ at timestep $t-1$, and $\tilde{\beta}_t$ denotes the variance calculated using the hyperparameters $\beta_t$, $\hat{\alpha}_t$, and $\hat{\alpha}_{t-1}$.

\subsection{Model details}
We employ a two-branch network architecture to model human interactions, where each branch processes the actions of one individual and information sharing occurs between the branches. Specifically, $x_a^t$ and $x_b^t$ are first processed through a motion embedding layer and sequence position encoding to generate initial hidden states $h^0_a$ and $h^0_b$. These states are then fed into a two-branch transformer network with shared weights, composed of $N$ transformer blocks. Within each block, self-attention (SA) and cross-attention (CA) mechanisms enable intra-agent and inter-agent information exchange, respectively. For the n-th transformer block in agent $a$'s branch where $n \in [1,N]$ :

The Self-Attention Block processes its own hidden state $h_a^{n-1}$ to capture intra-agent dependencies. The query $Q_a^{\text{SA}}$, key $K_a^{\text{SA}}$, and value $V_a^{\text{SA}}$ matrices are derived from $h_a^{n-1}$ as:

\begin{equation}
    \footnotesize
    Q_a^{\text{SA}} = h_a^{n-1} W_Q^{\text{SA}}, \quad K_a^{\text{SA}} = h_a^{n-1} W_K^{\text{SA}}, \quad V_a^{\text{SA}} = h_a^{n-1} W_V^{\text{SA}}
\end{equation}

where $W_Q^{\text{SA}}, W_K^{\text{SA}}, W_V^{\text{SA}}$ are trainable weights. The self-attention output is calculated as: 

\begin{equation}
    \text{SA} (h_a^{n-1}) = \text{Softmax}\left ( \frac{Q_a^{\text{SA}} (K_a^{\text{SA}})^T}{\sqrt{C}}\right) V_a^{\text{SA}}
\end{equation}

where $C$ is the number of channels in the attention layer. Then a Cross-Attention Block facilitates inter-agent information exchange. For agent a, the query matrix $Q_a^{\text{CA}}$ is derived from $h_a^{n-1}$, while the key $K_a^{\text{CA}}$ and value $V_a^{\text{CA}}$ matrices come from $h_b^{n-1}$:

\begin{equation}
    \footnotesize
    Q_a^{\text{CA}} = h_a^{n-1} W_Q^{\text{CA}}, \quad K_a^{\text{CA}} = h_b^{n-1} W_K^{\text{CA}}, \quad V_a^{\text{CA}} = h_b^{n-1} W_V^{\text{CA}}
\end{equation}

The cross-attention output for agent a is:

\begin{equation}
    \text{CA} (h_a^{n-1}, h_b^{n-1}) = \text{Softmax}\left (\frac{Q_a^{\text{CA}} (K_a^{\text{CA}})^T}{\sqrt{C}}\right) V_a^{\text{CA}}
\end{equation}

A symmetric calculation for agent b, $\text{SA} (h_b^{n-1}), \text{CA} (h_b^{n-1}, h_a^{n-1})$, swaps the roles of $h_a^{n-1}$ and $h_b^{n-1}$.The weight matrices $W_Q^{\text{SA}}, W_K^{\text{SA}}, W_V^{\text{SA}}$ and $W_Q^{\text{CA}}, W_K^{\text{CA}}, W_V^{\text{CA}}$ are shared across both branches. 
At the end of each block, the outputs of the SA and CA blocks are combined with residual connections and layer normalization, for agent a:

\begin{equation}
    h_a^{n} = \text{LayerNorm}\left (h_a^{n-1} + \text{SA} (h_a^{n-1}) + \text{CA} (h_a^{n-1}, h_b^{n-1})\right)
\end{equation}

This integrated hidden state $h_a^{n}$ is then fed into subsequent transformer layers.
The weight-sharing symmetry ensures balanced processing of inter-agent interactions, reducing model parameters while improving generalization capabilities.

ControlNet is a trainable copy of the $N$ transformer blocks of the diffusion model, they share common inputs: $h^0_a$, $h^0_b$, t, and $F_{\text{img}}$. Additionally, it incorporates text features $F_{\text{text}}$ encoded by CLIP. For each trained transformer block, the computation is defined as:
$
    h_i = \mathcal{T} (h^{i-1}, F_{\text{img}}; \Theta)
$, where $\Theta$ denotes the frozen parameters of the block.

The trainable copy of the model connects to the original model via zero linear layers. The output of the controlled diffusion network is therefore:

\begin{equation}
    \footnotesize
    h_i^c = \mathcal{T}(h^{i-1}, F_{\text{img}}; \Theta) + \mathcal{Z}\left( \mathcal{T}\left( x + \mathcal{Z}(F_{\text{text}}; \Theta_{z1}), F_{\text {img}}; \Theta_c \right); \Theta_{z2} \right)
\end{equation}

Here $\mathcal{T}$ represents the original model block and $\mathcal{Z}$ denotes the zero linear layers. At the start of training, the zero layers output zeros, so $h_i^c = \mathcal{T} (h_{i-1}^c; \Theta)$ matches the base model. As training proceeds, the zero layers gradually inject conditional signals.

\subsection{VLM annotator details}

We further provide the details of VLM Annotation in Tab.~\ref{tab:prompt}. We also provide several generated textual descriptions and contact pairs in Fig.CameraReady~\ref{fig:inthewild}; the text is well aligned with the images and supplies semantic guidance for human mesh reconstruction.

\begin{table*}[!ht]
    \vspace{-2ex}
    \caption{Detailed prompting example for VLM Annotator.}
    \centering
    \small
    \begin{tabularx}{\linewidth}{X}
    \toprule
    Prompting Example \\
    \midrule
    Given the image sequence of two human interaction, generate 0, 1 or more joint-joint contact pair(s) according to the following background information, rules, and examples. Joint-joint contact pair should exactly reflect the human interaction shown in the image sequence. \\
    
    \lbrack Start of background Information\rbrack\\
    Human has JOINTS: \lbrack `pelvis', `left\_hip', `right\_hip', `left\_knee', `right\_knee', `left\_ankle', `right\_ankle', `left\_foot', `right\_foot', `neck', `left\_collar', `right\_collar', `head', `left\_shoulder', `right\_shoulder', `left\_elbow', `right\_elbow', `left\_wrist', `right\_wrist' \rbrack.\\
    
    \lbrack End of background Information\rbrack \\
    
    \lbrack Start of rules\rbrack \\
    1.Each joint-joint pair should be formatted into \{JOINT, JOINT, TIME-STEP, TIME-STEP\}. JOINT should be replaced by JOINT in the background information. IMPORTANT: The first JOINT belongs to person 1, and the second JOINT belongs to person 2. Each joint-joint pair represents a contact of a joint of person 1 and a joint of person 2. The first TIME-STEP is the start frame number of contact, and the second TIME-STEP is the end frame number of contact. \\
    2.Use one sentence to describe what action person 1 do and one sentence to describe what action person 2 do according to the image sequence. IMPORTANT: the sentence starts from `text 1:' describing the action of person 1 from the perspective of person 1 and the sentence starts from `text 2:' describing the action of person 2 from the perspective of person 2. Sentences should NOT contain words like `person 1' or `person 2', use `a person' to refer to himself in the sentence and `others' to refer to others. IMPORTANT: the sentence should be align with the joint-joint contact pair. IMPORTANT: the order of person 1 and person 2 should be the same in different joint-joint contact pair of the same image sequence. \\
    3.IMPORTANT: Do NOT add explanations for the joint-joint contact pair.\\
    \lbrack End of rules\rbrack \\
    \lbrack Start of an example\rbrack \\
    \lbrack Start of sentences\rbrack\\
    Text 1: a person dance with others holding his left hand with the other's right hand, putting his right hand on the other's waist, and his shoulder being touched.\\
    Text 2: a person dance with other holding her right hand with the other's left hand, with her waist being embraced, placing her left hand on the other's shoulder.\\
    \lbrack End of sentences \rbrack\\
    \lbrack Start of joint-joint contact pair(s)\rbrack\\
    \{left\_wrist, right\_wrist, 11, 15\} \\
    \{right\_wrist, left\_hip, 14, 15\} \\
    \{right\_shoulder, left\_wrist, 9, 15\}\\
    \lbrack End of joint-joint contact pair(s)\rbrack \\
    \lbrack End of an example\rbrack \\
    \bottomrule
    \end{tabularx}
    \label{tab:prompt}
    \vspace{-2ex}
\end{table*}

\section{Additional Experiments}
\label{sec:add_exp}

\subsection{Ablation on Geometry Optimizer}

Geometry Optimizer focuses on processing 3D joint positions to provide geometric guidance information. To validate the effectiveness of our encoding layer design for the auxiliary model, we conducted an ablation study by implementing the motion embedding layer with either STGCN or a Linear layer. The results are presented in Tab.~\ref{tab:ablation-geo}.

\begin{table}[!ht]
\caption{Ablation studies on the impact of motion embedding layer.}
\centering
\small
\resizebox{\linewidth}{!}{
\begin{tabular}{l|ccccc}
\toprule
Embedding Layer & $\downarrow$R-MPJPE & $\downarrow$G-MPJPE & $\downarrow$Int & $\downarrow$MPJPE & $\downarrow$PA-MPJPE \\
\toprule
Linear & 102.3 & 110.0 & 84.9 & 81.9 & 66.8 \\
STGCN & \textbf{81.7} & \textbf{93.2} & \textbf{62.5} & \textbf{60.8} & \textbf{47.8} \\
\bottomrule
\end{tabular}}
\label{tab:ablation-geo}
\end{table}

The Geometry Optimizer that uses STGCN to encode 3D joint positions exhibits higher accuracy than the one using Linear. It successfully captures the 3D positional relationships of interacting humans and outperforms Motion Infiller in all metrics related solely to 3D joint positions. This indicates that it can effectively provide correct guidance information.

\subsection{Additional experiments results on Hi4D}
We further partition Hi4D into subsets by action label to assess performance across interaction categories. Tab.~\ref{tab:hi4d-subset} presents our method's improvements over CloseInt across different subsets. Notably, our approach achieves the most significant gains on actions such as handshake, high-five, and kiss. In these actions, human behavioral patterns are relatively uniform, and occlusion levels are moderate. The model synthesizes plausible poses by integrating textual descriptions generated by VLM Annotator, while simultaneously mitigating mesh interpenetration issues and refining contact relationships.
However, the method shows smaller gains on complex actions such as dancing and fighting. These activities involve intricate limb interactions and ambiguous joint-depth relationships, which can slightly undermine VLM annotation consistency and the precision of geometric guidance. Nonetheless, our method still outperforms the baseline.

\begin{table*}[!ht]
\caption{Comparison of CloseInt and our method, CloseInt/Ours (Improvement), across different actions on Hi4D. }
\centering
\small
\resizebox{\linewidth}{!}{
\begin{tabular}{l|c c c c c}
\toprule
Action & handshake & high-five & kiss & dance & fight \\
\midrule
$\downarrow$R-MPJPE & 78.0/65.8 (20.0) & 60.5/53.5 (19.2) & 81.7/67.7 (20.9) & 96.4/87.6 (9.4) & 110.3/100.3 (8.2) \\
$\downarrow$G-MPJPE & 93.2/72.5 (23.8) & 84.9/70.9 (19.7) & 98.9/79.7 (19.6) & 109.2/97.9 (9.4) & 131.4/120.1 (6.5) \\
$\downarrow$Int & 36.9/31.1 (15.6) & 26.7/25.1 (5.9) & 33.3/23.0 (63.9) & 39.9/32.4 (18.8) & 46.7/41.1 (11.9) \\
$\downarrow$Pen & 194.8/71.9 (63.1) & 107.4/50.1 (53.3) & 15409.3/5570.1 (63.9) & 5477.1/2455.0 (55.2) & 636.6/226.2 (64.5) \\
\bottomrule
\end{tabular}}
\label{tab:hi4d-subset}
\end{table*}

\subsection{Cross dataset evaluation}

We also report both intra-domain and cross-domain results. SocialMirror outperforms prior methods in all settings. In our experiments, we observed that when not trained on the dataset, CloseInt may erroneously separate characters that should be in close contact. This results in the absence of even minor intended penetrations (e.g., slight mesh intersection between a palm and another person), leading to a relatively low penetration error—though this is not indicative of a good reconstruction outcome. After training on the dataset, CloseInt's errors in character placement are reduced, but it correspondingly exhibits more interpenetration, which explains why the penetration loss increases post-training. Our method, in both scenarios, produces more accurate relative positions of characters (as reflected in RE and Int) while ensuring less interpenetration, demonstrating the positive effect of the proposed method in reducing interpenetration.

\begin{table*}[t]
    \centering
    \caption{Cross Dataset Evaluation on Hi4D and 3DPW.}
    \label{tab:Cross Dataset Evaluation on Hi4D and 3DPW.}
    \small
    \resizebox{\linewidth}{!}{
    \begin{tabular}{l|ccccccc|ccccccc}
    \toprule
    \multirow{2}{*}{Method} & \multicolumn{7}{c|}{Hi4D} & \multicolumn{7}{c}{3DPW} \\
    \cmidrule (lr){2 - 8} \cmidrule (l){9 - 15}
    & $\downarrow$RE & $\downarrow$GE & $\downarrow$Int & $\downarrow$Pen & $\downarrow$MPJPE & $\downarrow$PA. & $\downarrow$VPE & $\downarrow$RE & $\downarrow$GE & $\downarrow$Int & $\downarrow$Pen & $\downarrow$MPJPE & $\downarrow$PA. & $\downarrow$VPE \\
    \midrule
    CloseInt & 99.0 & 114.9 & 81.4 & 3947.6  & 63.1 & \textbf{47.5} & \textbf{76.4} & 135.7 & \textbf{159.1} & 95.5 & 342.7  & 79.9 & 52.9 & 95.1 \\
    Ours & \textbf{83.6} & \textbf{95.2} & \textbf{68.5} & \textbf{2380.5}  & \textbf{62.2} & \textbf{47.5} & 79.3 & \textbf{104.8} & 162.7 & \textbf{89.9} & \textbf{109.7}  & \textbf{65.1} & \textbf{49.0} & \textbf{79.7} \\
    \midrule
    CloseInt (Eval. Only) & 181.1 & 232.1 & 182.7 & \textbf{1973.8} & 109.1 & \textbf{62.5} & 132.0 & 194.4 & 340.2 & 128.4 & \textbf{101.6}  & 88.6 & 63.6 & 110.7 \\
    Ours (Eval. Only) & \textbf{165.2} & \textbf{184.1} & \textbf{153.0} & 2380.3  & \textbf{105.2} & 63.6 & \textbf{129.4} & \textbf{174.7} & \textbf{307.4} & \textbf{125.4} & 109.7  & \textbf{87.5} & \textbf{63.3} & \textbf{109.8} \\
    \bottomrule
    \end{tabular}
    }
\end{table*}

\subsection{Results on Harmony4D}

For completeness, we also conducted training experiments on the Harmony4D dataset, which further confirms the effectiveness of our approach. Specifically, our method achieves significant improvements in interaction-related metrics: it yields decreases of 8.2\%, 3.5\%, and 3.2\% in RE, GE, and Int, respectively. Meanwhile, it maintains nearly unchanged performance on single-person reconstruction metrics (i.e., MPJPE, PA, and VPE). This result demonstrates the robust capability of our method in capturing human interaction relationships.

\begin{table}[ht]
    \centering
    \caption{Comparisons on Harmony4D.}
    \label{tab:Comparisons on Harmony4D (train on).}
    \small
    \resizebox{\linewidth}{!}{
    \begin{tabular}{l|cccccccc} 
    \toprule
    Method & $\downarrow$RE & $\downarrow$GE & $\downarrow$Int & $\downarrow$Pen & $\downarrow$MPJPE & $\downarrow$PA. & $\downarrow$VPE \\
    \midrule
    CloseInt & 134.8 & 297.5 & 182.5 & 482.6 & 70.2 & 38.6 & 82.6 \\
    Ours & \textbf{123.8} & \textbf{287.2} & \textbf{176.8} & \textbf{480.3} & \textbf{69.8} & \textbf{39.7} & \textbf{80.8} \\
    \bottomrule
    \end{tabular}
    }
\end{table}

\subsection{Additional visualization results}

We present additional in-the-wild reconstructions in Fig.~\ref{fig:inthewild}; the supplementary video includes further comparisons and demonstrations.

\section{Discussions}
\label{sec:discussions}

\subsection{Reconstruction under VLM Limitations}
\begin{figure*}
    \centering
    \vspace{10ex}
    \includegraphics[width=\linewidth]{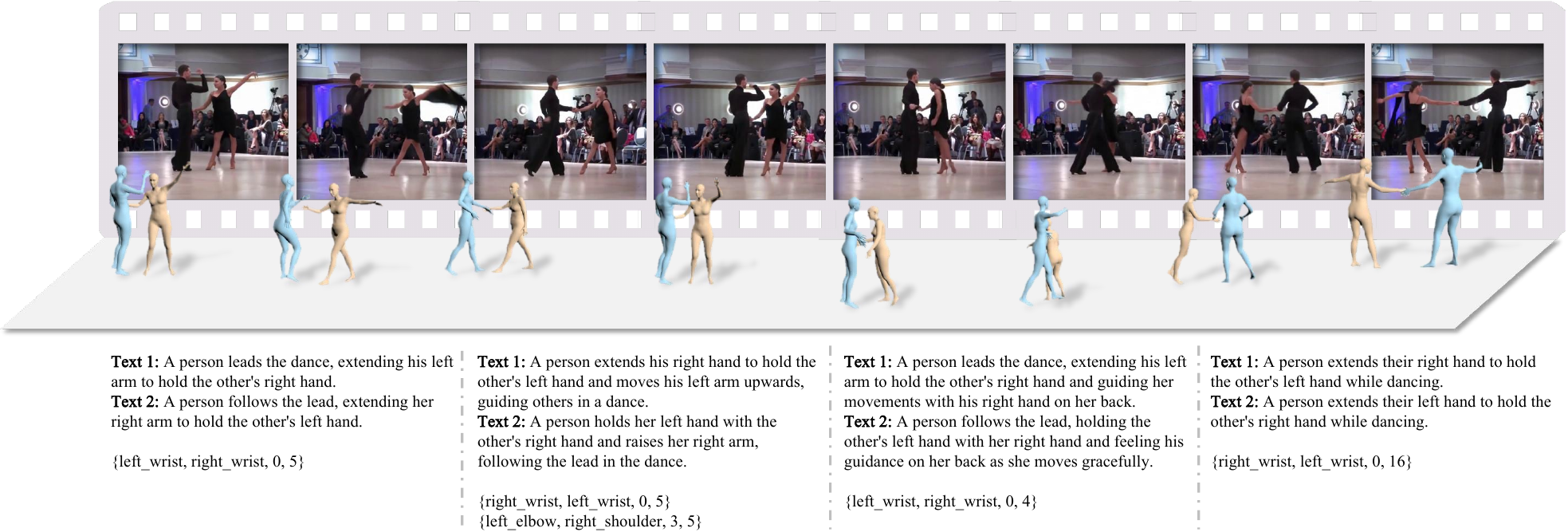}
    \includegraphics[width=\linewidth]{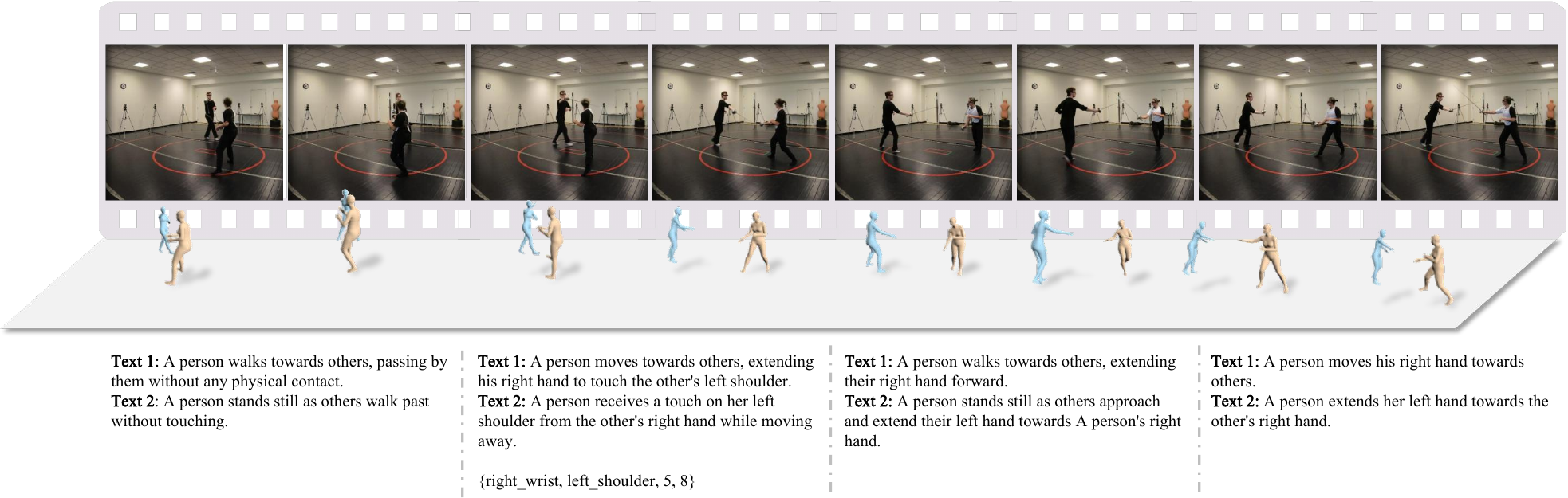}
    \includegraphics[width=\linewidth]{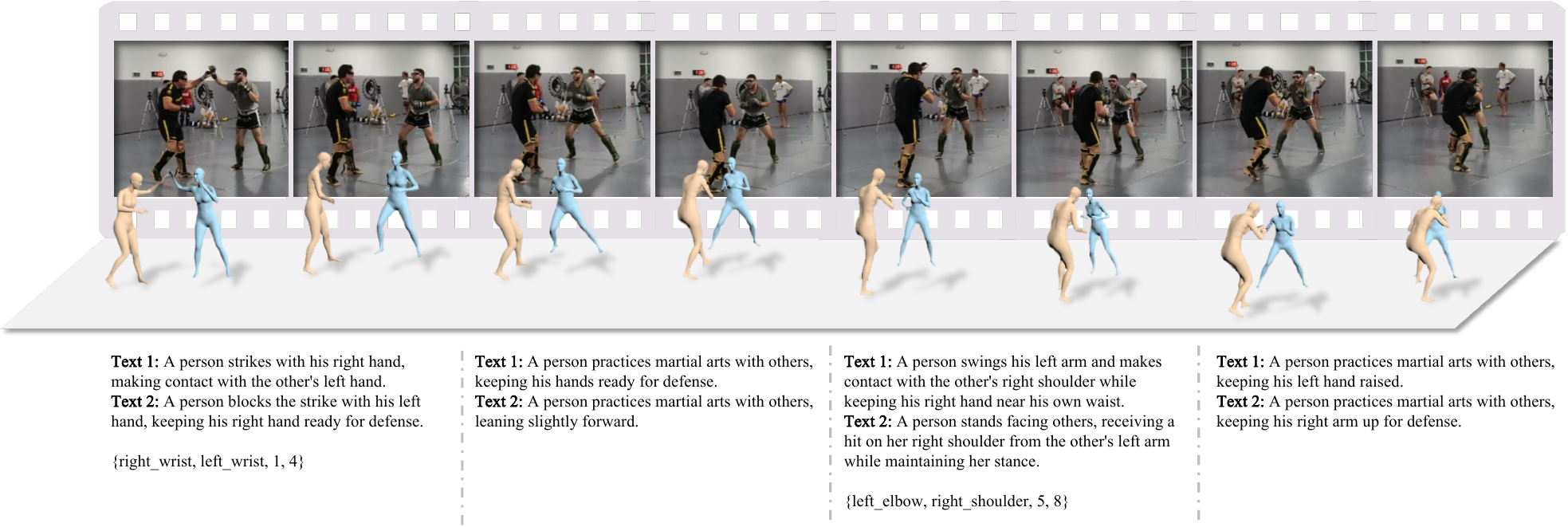}
    \caption{Visualization results on in-the-wild data.}
    \label{fig:inthewild}
    \vspace{10ex}
\end{figure*}

\begin{figure*}
\vspace{-4ex}
    \centering
    \includegraphics[width=\linewidth]{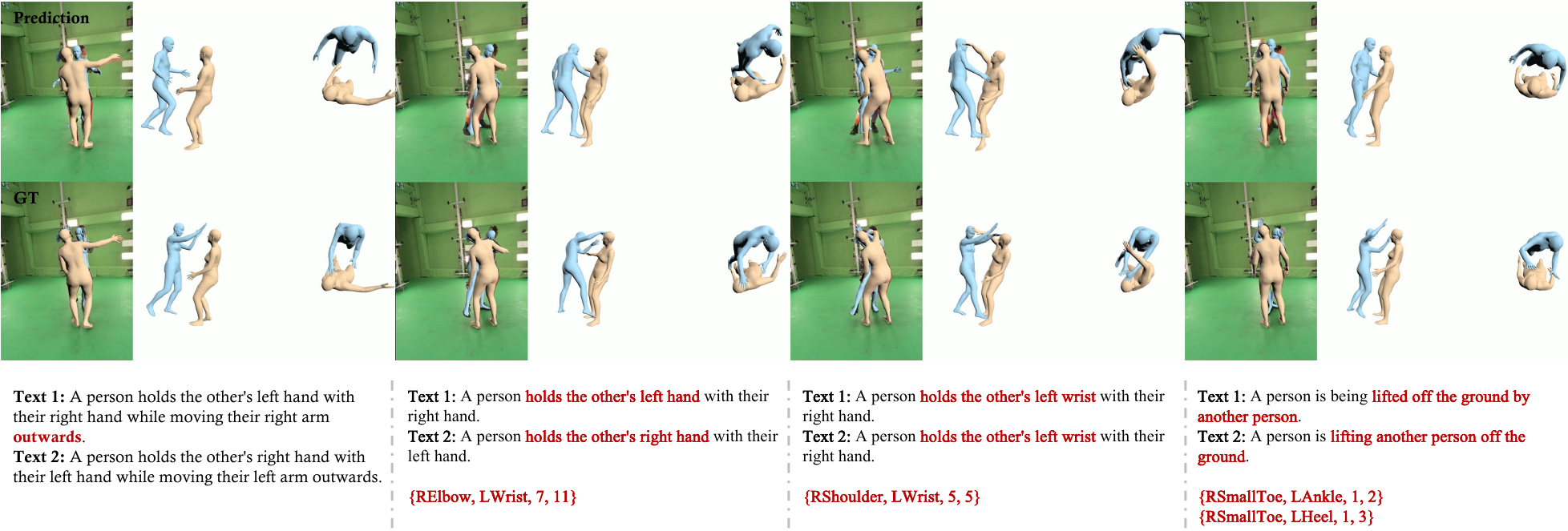}
    \vspace{-5ex}
    \caption{VLM Annotator failed to describe human interaction.}
    \label{fig:vlm_fail}
    \vspace{-2ex}
\end{figure*}

Based on our user study, the text descriptions generated by the VLM are, on average, superior to those produced by human annotators. As shown in Fig.\ref{fig:inthewild}, VLM annotations can capture not only macroscopic actions but also fine-grained contact relationships between specific joints (e.g., “A person leads the dance, extending his left arm to hold the other's right hand and guiding her movements with his right hand on her back”), whereas a human annotator might simply describe it as "two people dancing ballroom dance. 
While VLM Annotator demonstrates satisfactory performance in describing human interaction under most circumstances, its accuracy tends to decline when confronted with complex limb interactions, affecting the precision of both textual descriptions and contact pair annotations. By prioritizing visual feature extraction over textual inputs, our proposed method maintains reconstruction fidelity even when text-image alignment is compromised. As illustrated in Fig.\ref{fig:vlm_fail}, despite VLM Annotator's failure to correctly identify the human action, our approach successfully reconstructs accurate motion patterns by leveraging visual information.

\begin{figure*}
    \centering
    \includegraphics[width=\linewidth]{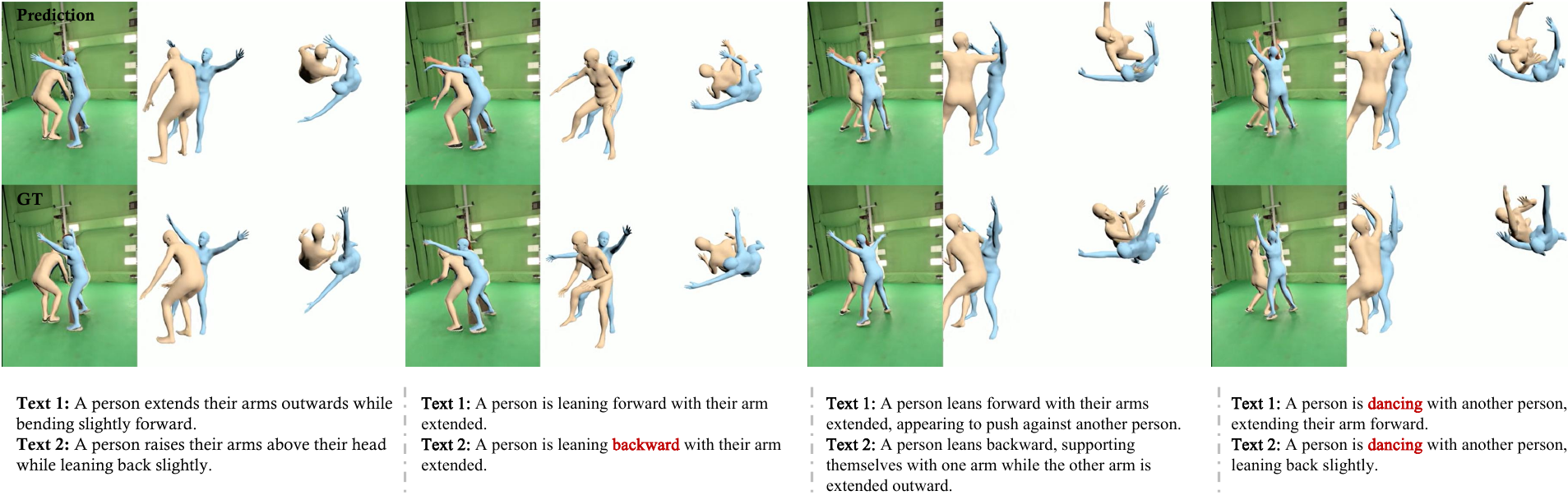}
    \vspace{-5ex}
    \caption{Challenging case with prolonged, severe occlusions.}
    \label{fig:failure_case}
    \vspace{-2ex}
\end{figure*}

\subsection{Failure cases}
Our approach remains limited under prolonged, severe occlusions. Fig.~\ref{fig:failure_case} shows a case where both visual and semantic cues are unreliable. Inaccurate text and contact predictions from the VLM annotator then propagate misleading guidance and large reconstruction errors. This observation underscores the necessity of complementary mechanisms to handle extreme occlusion scenarios in future work.

\subsection{The effect of semantic information on limited contact scenarios}

Even when contact is absent, the VLM can still produce high-level scene descriptions (e.g., two people stand and face each other), which are encoded as semantic features. These provide contextual cues about interaction and spatial layout beyond direct contact information. In addition, our approach does not rely solely on contact labels.
The semantic features guide the Motion Infiller to infer plausible poses for ambiguous regions, and the Temporal Motion Refiner and geometric constraints based on 3D joint prediction from the Auxiliary Module ensure motion smoothness and spatial plausibility. Table~\ref{tab:hi4d-subset} further shows gains in interaction metrics even for actions with mild occlusion and limited contact.


\subsection{Limitations and future works}

Our current pipeline targets two-person interaction. For reconstructing interactions involving more participants, further improvements to the network architecture and annotation protocols are required.

Improving the reliability of semantic guidance is another important direction for future work. Promising steps include estimating confidence from the VLM annotator, adaptively reweighting text conditioning when captions are uncertain, and explicitly checking semantic-visual agreement before feeding language into reconstruction.

\section{The Use of Large Language Models (LLMs)} 
\label{sec:use_llm}
We declare that vision-language models (VLMs) in this paper are used primarily as a VLM Annotator to produce textual descriptions of interactions in image sequences and spatio-temporal joint contact pairs. LLMs are used only for light text polishing and grammar fixes. The research approach, core ideas, reasoning, and conclusions remain the authors' own work. All VLM/LLM-assisted content generation is documented together with how and where it was applied.


\end{document}